\documentclass[lettersize,journal]{IEEEtran}
\usepackage{amsmath,amsfonts}
\pdfoutput=1
\usepackage{algorithmic}
\usepackage{array}
\usepackage[caption=false,font=normalsize,labelfont=sf,textfont=sf]{subfig}
\usepackage{textcomp}
\usepackage{stfloats}
\usepackage{url}
\usepackage{booktabs}
\usepackage{verbatim}
\usepackage{graphicx}
\def\BibTeX{{\rm B\kern-.05em{\sc i\kern-.025em b}\kern-.08em
    T\kern-.1667em\lower.7ex\hbox{E}\kern-.125emX}}
\usepackage{balance}
\usepackage{multirow}
\usepackage[table,xcdraw]{xcolor}
\usepackage[normalem]{ulem}
\useunder{\uline}{\ul}{}
\begin{document}

\title{Shift-Window Meets Dual Attention: A Multi-Model Architecture for Specular Highlight Removal}

\author{
Tianci Huo,
Lingfeng Qi,
Yuhan Chen,
Qihong Xue,
Jinyuan Shao,
Hai Yu,
Jie Li,
Zhanhua Zhang, 
and Guofa Li,~\IEEEmembership{Senior Member,~IEEE}

\thanks{Tianci Huo, Lingfeng Qi, Yuhan Chen, Qihong Xue, Jinyuan Shao, Hai Yu, Jie Li, and Guofa Li, are with the College of Mechanical and Vehicle Engineering, Chongqing University, Chongqing, 400044, China.}
\thanks{Zhanhua Zhang is with the College of Computer Science and Technology, Zhejiang University, Hangzhou, 310058, China and Artificial Intelligence Center, Geely Automotive Research Institute, Ningbo, 315300, China.}
}

\maketitle
\begin{abstract}
Downstream tasks in visual multimedia and computer vision, such as image fusion and image segmentation, are severely affected by specular reflections. However, existing methods exhibit limited texture consistency and struggle to balance inference speed with highlight removal accuracy. To resolve these challenges, this paper proposes a multi-model architecture for specular highlight removal (MM-SHR). We employ convolution operations to extract local details in the shallow layers of MM-SHR, and utilize attention mechanism to capture global features in the deep layers, ensuring both operational efficiency and removal accuracy. To model long-range dependencies without compromising computational complexity, we propose Omnidirectional Attention Integration Block and Adaptive Region-Aware Hybrid-Domain Dual Attention Convolutional Network, which leverage pixel-shifting and window-dividing operations at the raw features to achieve specular highlight removal in a coarse-to-fine manner. Moreover, frequency processing at different feature scales is incorporated in the proposed modules to incrementally enhance spatial perception of highlight features. This facilitates capturing global illumination features while preserving the overall brightness of the original image by avoiding over-processing. Extensive experiments show that MM-SHR exceeds the state-of-the-art methods investigated in evaluation metrics. The implementation will be made publicly available at https://github.com/Htcicv/MM-SHR.
\end{abstract}


\section{Introduction}

\IEEEPARstart{S}{pecular} highlight is a localized highlight phenomenon caused by specular reflection on smooth surfaces under the action of light, and this localized highlight often reduces the visual consistency of images and videos and destroys the detailed texture information, which poses great challenges to computer vision tasks, including image segmentation, NeRF Rendering, and object detection~\cite{RN49}. 
Therefore, the efficient removal of specular highlight from images is of great importance for the development and application of multimedia vision technology.
To solve this problem, researchers have proposed various methods of specular highlight removal. 
A classic approach is based on the HIS color space~\cite{RN01}, where researchers achieve specular component separation by matching the saturation of specular pixels to the value of diffuse pixels of the same diffuse chromaticity. 
Another widely used method is the method~\cite{RN06,RN13,RN14,RN15} based on the dichromatic reflection model~\cite{RN02}, where researchers have achieved the removal of specular highlight from an image by separating the reflective component of the image into two parts: diffuse and specular reflection. 
In addition to methods based on HIS and dichromatic reflection model, filters~\cite{RN07}, polarization characterization~\cite{RN08}~\cite{RN09}~\cite{RN12} and principal component analysis (PCA)~\cite{RN10} have been newly applied to the field of specular highlight removal.
Despite their contributions, these traditional methods generally depend on restrictive assumptions regarding surface properties, lighting conditions, or material priors, which undermines their applicability to complex real-world scenes. 

With the rapid development of deep learning~\cite{RN29,RN30} and high-quality datasets~\cite{RN37,RN41,RN43}, a variety of deep learning-based specular highlight removal methods have emerged, which can be broadly categorized into multi-image approaches~\cite{RN16,RN17,RN20} and single-image approaches~\cite{RN26,RN27,RN33,RN34}.
Multi-image methods often leverage highlight masks or multi-view information to improve accuracy but require auxiliary data, limiting their applicability.
Single-image methods are more practical, but face inherent trade-offs: CNN-based networks lack long-range feature modeling, while Transformer-based architectures incur high computational costs~\cite{RN34}, making real-time deployment challenging. 
Furthermore, many existing UNet-based designs employ homogeneous encoder–decoder architectures—constructed entirely with CNNs or Transformers across all scales—which fail to exploit their complementary strengths in capturing local details and global context. 
These limitations motivate our proposed method.

To solve the above challenges while ensuring both real-time performance and high precision for practical applications, we propose a novel end-to-end network for specular highlight removal.
Our approach combines shallow and deep network components to optimize feature extraction and model efficiency. 
In particular, to enhance the shallow network’s ability to model long-range dependencies, we incorporate the window partitioning mechanism from the Swin Transformer~\cite{RN11} into its architecture, enabling it to capture global context more effectively while preserving local feature details.
Furthermore, to overcome the inherent limitation of isolated window processing in existing window-based models, we design a novel shallow-level cross-window modeling mechanism. 
In this mechanism, feature maps are first divided into non-overlapping windows, which are then shifted in multiple directions.
For each window, we perform joint global–local feature extraction guided by frequency-domain information, allowing the shallow network to capture long-range dependencies without sacrificing fine-grained local details. 
This design, for the first time in the field of specular highlight removal, eliminates the isolation problem between windows while enabling cross-window global modeling at the shallow stage of the network.
Our primary contributions are summarized as follows:
\begin{enumerate}
	\item We develop a multi-model architecture for specular highlight removal (MM-SHR). This network adopts a hierarchical structure that combines CNN for detail extraction and Transformer for global feature integration, leveraging the strengths of both to efficiently and effectively remove specular highlight. Meanwhile, we leverage the frequency-domain information, which includes global illumination characteristics, to guide the extraction of spatial highlight features across across multiple scales.
	\item To enhance the CNN's ability for capturing long-range dependencies, we incorporate  omni-directional pixel shifting and window dividing operations, which segments the raw features into interrelated windows, into both the Omnidirectional Attention Integration Block (OAIBlock) and the Adaptive Region-Aware Hybrid-Domain Dual Attention Convolutional network (HDDAConv), which distinguishes itself from OAIBlock by decoupling these operations. This allows MM-SHR to more fully capture the dependencies of long-range information across multiple scales.
	\item Extensive experiments conducted on the SHIQ (Specular Highlight Image Quadruples), PSD (Paired Specular-Diffuse), and NSH (Natural Specular Highlight) datasets demonstrate that MM-SHR outperforms state-of-the-art methods both quantitatively and qualitatively. Additional ablation studies confirm the individual effectiveness of each proposed module, providing a compelling and generalizable solution for specular highlight removal tasks in real-world imaging systems.
\end{enumerate}

\section{Related Work}
\subsection{Multi-image Methods}
Multi-image-based techniques utilize auxiliary information such as multiple viewpoints or externally generated highlight masks to facilitate specular highlight removal. 
Among these methods, Feng et al.~\cite{RN16}  introduced an approach based on the dichromatic reflection model, utilizing full variational optimization within a light field camera to eliminate highlight. 
Fu et al.~\cite{RN17} introduced a stage-wise approach for specular reflection removal. 
Jin et al.~\cite{RN18} proposed a two-step learning framework that first employs a constrained loss function based on prior knowledge of shaded and highlight-free images to estimate an initial reflectance layer. This is followed by an S-aware network that further refines the reflectance layer. 
Ultimately, their model classifies and separates the highlight component from the input image. Researchers~\cite{RN19,RN24,RN25} have explored the use of highlight masks to guide specular highlight removal. 
Wu et al.~\cite{RN20} developed a UNet-Transformer network that detects and removes highlight by first training a UNet-based highlight detection module and then using the generated mask to guide the removal process. 
Hu et al.~\cite{RN21} applied the NMF method to extract highlight region mask and separately trained generator networks for the luminance and full channels. 
Their approach produced high-quality results free of color distortion and highlight residues.
Despite their theoretical soundness, mask-dependent methods often face limitations in real-world deployment because highlight mask annotations are often unavailable or inaccurate. 
Recently, Ha et al.~\cite{RN22} addressed the challenge of accurately separating diffuse and specular reflection in high dynamic range images by introducing the Temporal Dark Prior (TDP) as a pseudo-diffuse reflection. 
At the same time, they leveraged the periodic variation characteristics of AC-powered light sources as constraints, offering a novel approach to highlight removal. 
For facial highlight removal, Su et al.~\cite{RN23} employed a nonlinear optimization approach based on multi-view face images and Lambertian consistency to effectively reduce highlight. 
Overall, while multi-image-based specular highlight removal techniques have achieved significant success, their reliance on multiple images or auxiliary data limits their practical applicability.

\subsection{Single-Image methods}
Single-image specular highlight removal methods aim to eliminate highlights from a single image, offering broad applicability compared with multi-image approaches. 
Early works often combined traditional reflection models with handcrafted strategies, such as coarse-to-fine associative learning~\cite{RN26}, feature decomposition~\cite{RN27}, and color expansion mechanisms~\cite{RN28}. 
With the rapid advancement of deep learning, specular highlight removal methods~\cite{RN20}~\cite{RN31}~\cite{RN32} based on new network architectures such as GAN~\cite{RN29} and Transformer~\cite{RN30} have emerged continuously.
Among these, Anwer et al.~\cite{RN33} proposed SpecSeg, a segmentation-oriented network based on the UNet architecture, which serves as a strong baseline for subsequent highlight removal tasks.
Guo et al.~\cite{RN34} proposed a highlight removal network DHAN-SHR built on the Transformer framework.
However, the high computational and memory demands of Transformers result in slow inference speeds, making the model unsuitable for real-time applications in automated systems. 
Madessa et al.~\cite{RN36} focused on improving the efficiency of existing methods for removing highlight from images of transparent materials. 
They employed automatic semantic mask generation and partial convolution restoration techniques to effectively remove highlight regions, all while maintaining important local and semantic information in the image. 
Zhang et al.~\cite{RN37} recently introduced the NSH dataset for specular highlight removal, utilizing cross-polarization and the dichromatic reflection model to address the shortcomings of previous datasets such as SHIQ, SSHR, and PSD.
Overall, single-image methods provide greater robustness and practicality than multi-image approaches. 
Despite advances driven by deep learning, challenges remain: convolutional networks struggle with capturing long-range dependencies, while Transformers often require excessive computation. 
This trade-off between speed and removal quality motivates our work to develop more efficient architectures that better balance these factors and improve color and texture consistency.

\section{Methodology}
\subsection{Overall}
\begin{figure*}[!b]
	\includegraphics[width=\linewidth]{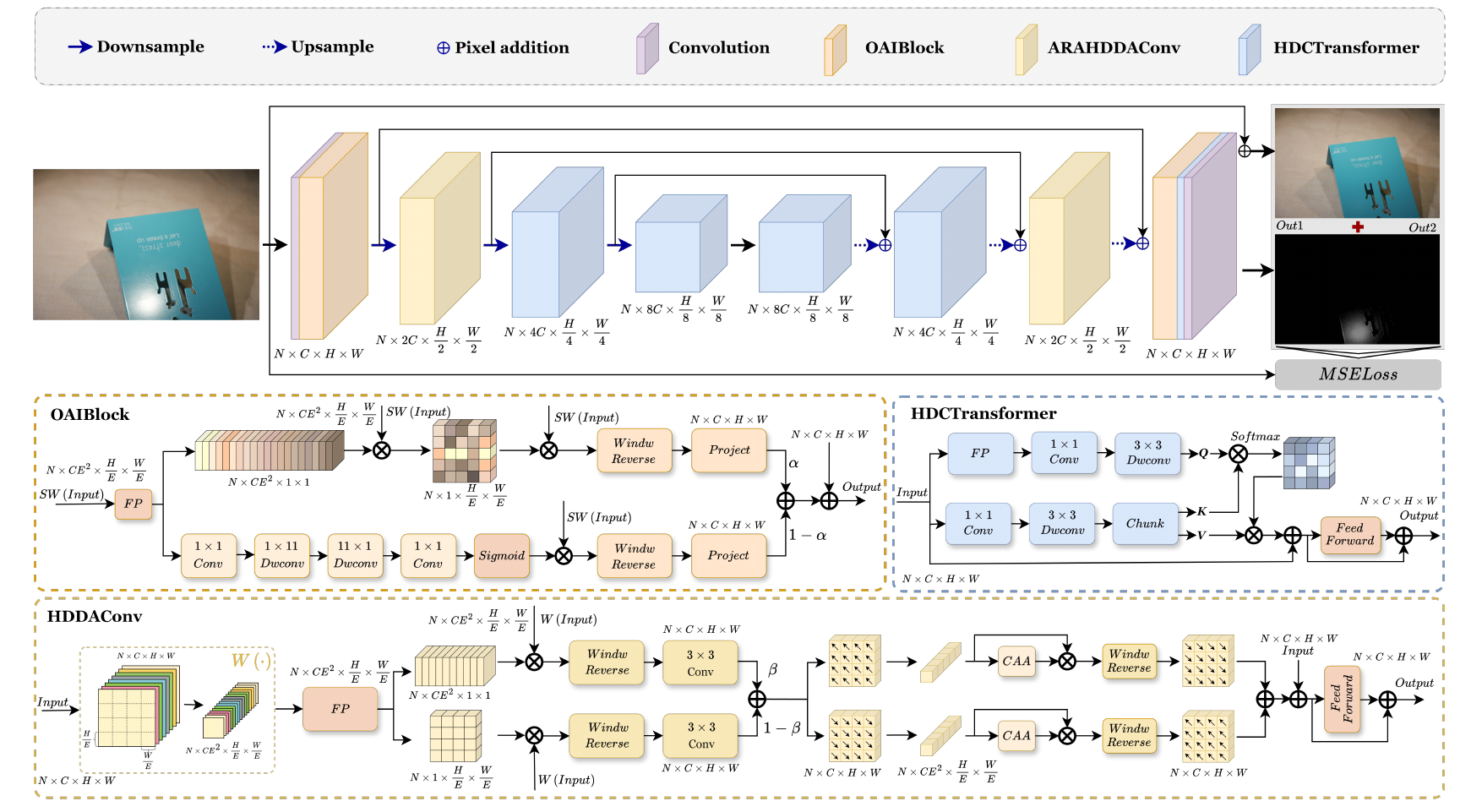}
	\caption{Overall architecture of Shift-Window Meets Dual Attention: A Multi-Model Architecture for Specular Highlight Removal}
	\label{Fig1}
\end{figure*}

The general architecture of MM-SHR, an end-to-end multi-model architecture with a single image as input, is shown in Fig.~\ref{Fig1}. 
Its purpose is to enhance the speed of inference and consistency of color texture for tasks involving the removal of specular highlight. 
The network constructs a multi-stage U-shaped encoder-decoder structure by ingeniously fusing the global feature capture advantage of Transformer with the local detail extraction capability of CNN. MM-SHR effectively captures and processes the global context and detailed texture in images at various scales by utilizing the synergy of residual concatenation, upsampling, and downsampling modules. The framework offers a dependable solution for specular highlight removal tasks by accurately separating and restoring features in highlight and non-highlight regions using pixel-by-pixel feature accumulation and multipath feature fusion strategies.

The framework proposed in this paper consists of three key modules: the Omnidirectional Attention Integration Block (OAIBlock), the Adaptive Region-Aware Hybrid-Domain Dual Attention Convolutional Network (HDDAConv), and the Hybrid-Domain Cross-Attention Transformer (HDCTransformer). 

\subsection{Omnidirectional Attention Integration Block}
Specular highlight often appear as localized highlight regions lacking detail, resulting in significant loss of local information. Relying solely on local detail features makes it challenging to reconstruct the original texture accurately. To address this, capturing broader contextual information becomes essential, especially at shallower network layers. While Transformer excel at capturing long-range dependencies, it introduce considerable memory overhead and slow down inference. To balance performance and efficiency, we propose an Omnidirectional Attention Integration Block (OAIBlock) based on convolutional operations. Positioned at the topmost layer of our MM-SHR network, the OAIBlock captures medium-range and long-range contextual information from high-resolution features. Simultaneously, it enhances local detail extraction and feature fusion without significantly increasing floating-point operations and inference time. This design ensures low-cost, high-quality extraction of low-level highlight features, laying a robust foundation for subsequent deep-level feature extraction and improving overall highlight removal performance.
OAIBlock mainly realizes the coupling of medium-range and long-range features through Omnidirectional Pixel Shifting and Window Dividing Operations (Shift-Window). Global-Local Attention (GLA) and Context Anchor Attention (CAA)~\cite{RN38} are also utilized to efficiently extract multi-scale contextual information. And the highlight features of multiple attentions are adaptively fused by a learnable parameter $\alpha$. The OAIBlock can be represented by Eq.~(\ref{eq1}):
%

\begin{equation}
\left\{
\begin{aligned}
    F &= SW(Input) \\
    OAIBlock(F) &= \alpha \times GLA(FP(F), F) \\
    &\quad + \left(1 - \alpha\right) \times CAA(FP(F), F)
\end{aligned}
\right.
	\label{eq1}
\end{equation}
where $Input\in R^{N\times C\times H\times W}$ represents the input features, $SW\left(\cdot\right)$ denotes the Omni-Directional Pixel Shifting and Window Dividing Operations (Shift-Window), $FP\left(\cdot\right)$ corresponds to the frequency processor, $GLA\left(\cdot\right)$ represents the Global-Local Attention mechanism, $CAA(\cdot)$~\cite{RN38} denotes the Contextual Anchor Attention mechanism.

\subsubsection{Shift-Window}
\begin{figure}[!h]
	\includegraphics[width=\linewidth]{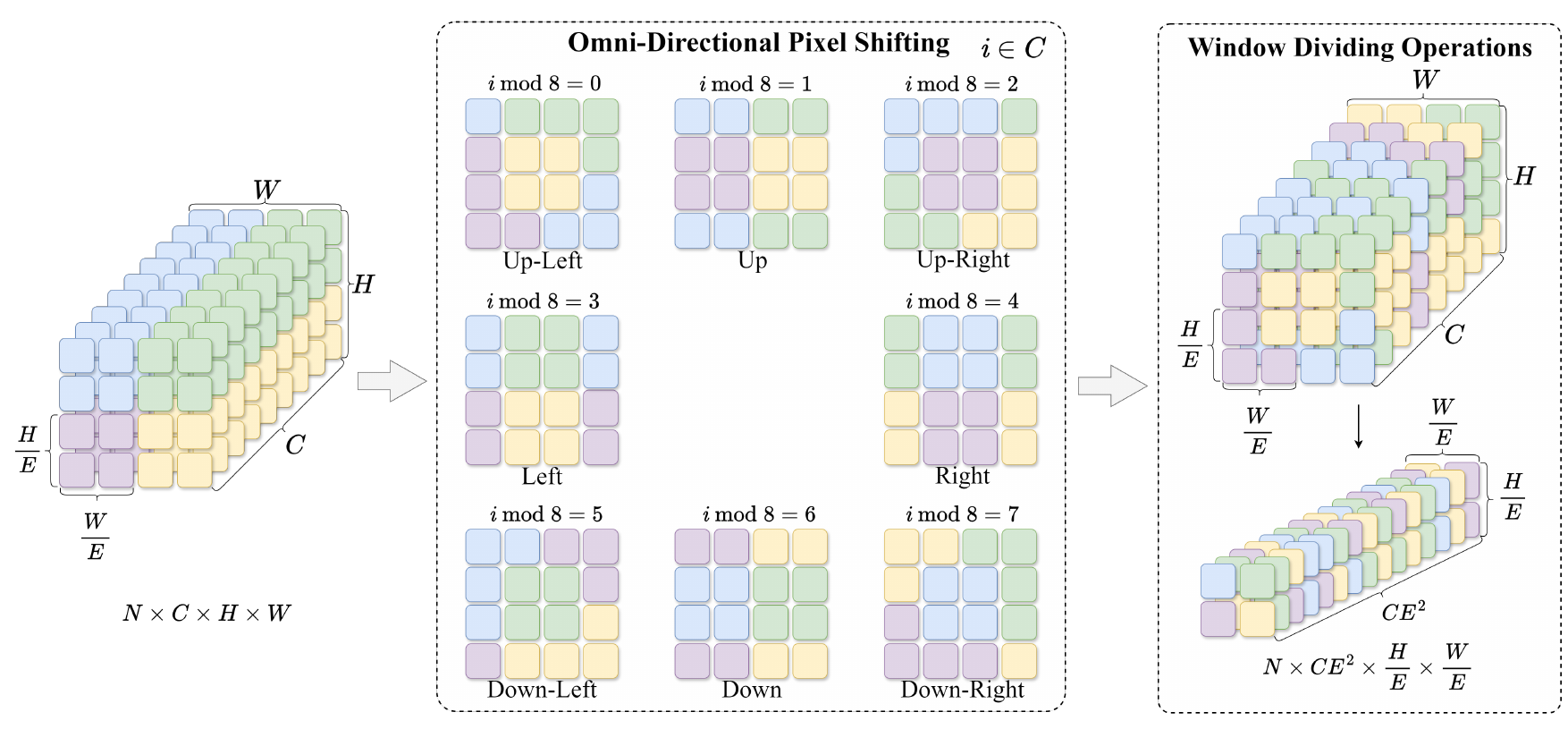}
	\caption{Diagram of the Shift-Window process}
	\label{Fig2}
\end{figure}

Specular highlight in images often obscure or eliminate fine details within localized regions. 
To restore these areas, information from the areas surrounding the specular highlights is needed as a reference. 
Some existing approaches~\cite{RN20}~\cite{RN34}~\cite{UDAformer} use Transformer-based architectures to capture long-range dependencies between highlight areas and their surrounding features. 
While effective, these methods are computationally expensive when applied to high-resolution features, leading to higher inference costs. 
Other approaches combine pixel shifting with window partitioning, but these are often implemented within Transformers and yield limited performance gains.
On the other hand, CNN-based architectures~\cite{RN17}~\cite{RN42} are efficient to train and run, but struggle with highlight removal due to their restricted receptive fields.
To overcome these limitations, we introduce $SW\left(\cdot\right)$ into CNNs without altering their receptive field. 
This preserves the efficiency of convolution while enabling effective modeling of long-range dependencies. The implementation is illustrated in Fig.~\ref{Fig2} and Eq.~(\ref{eq2}). 
\begin{equation}
\left\{
\begin{aligned}
    \mathcal{S}  &= [Up, Down, Left, Right, \mathrm{Left\text{-}Up}, \\
         &\mathrm{Right\text{-}Up}, \mathrm{Left\text{-}Down},\mathrm{Right\text{-}Down}]\\
    SW(Input) &= \sum_{\mathit{i}=1}^{C}\mathcal{D} (\mathcal{S}_{\mathit{i}\text{ mod 8}}(Input))
\end{aligned}
\right.
	\label{eq2}
\end{equation}

Given an input feature map $Input\in R^{N\times C\times H\times W}$, the $\mathit{i}\text{-th}$ channel in the channel dimension $C$ is cyclically shifted along the $\mathcal{S}_{\mathit{i}\text{ mod 8}}(\cdot)$ direction. 
To ensure complete window partitioning along both $H$ and $W$, the shifted feature maps are adaptively padded before division into windows of size $(H/E, W/E)$, where $E$ denotes the window expansion factor. 
The resulting output $F\in R^{N\times C E^2\times\frac{H}{E}\times\frac{W}{E}}$ simultaneously encodes cross-window semantic correlations and local geometric details.

\subsubsection{Frequency Processing}
Analyzing high-frequency components in the frequency domain allows the special highlight removal algorithm to effectively capture edges, fine details, and the global distribution of light in highlighted regions. This approach addresses the challenge of perceiving global highlight patterns, which is difficult when working solely in the spatial domain. To enhance feature representation, we apply a frequency-domain transformation, denoted as $FP\left(\cdot\right)$, to the input features $F$. This process improves the model’s ability to balance global highlight awareness with local detail preservation. The transformed features are then mapped back to the spatial domain for further processing. The specific implementation of $FP\left(\cdot\right)$ is presented in Eq.~(\ref{eq3}): 

\begin{equation}
	\begin{aligned}
		FP(F) &= Conv_{1\times1} \Big(Concat \Big(IFFT2D \Big(Conv_{1\times1}(BN\\ 
		&\quad (Re(FFT2D(F)))) + Re(FFT2D(F)) \Big), F \Big) \Big)
	\end{aligned}
	\label{eq3}
\end{equation}
where the operation $Concat\left(\cdot\right)$ combines the enhanced frequency domain features with the original spatial domain features, integrating both representations. $Re\left(FFT2D\left(\cdot\right)\right)$ extracts key frequency information, while $IFFT2D\left(\cdot\right)$ transforms the processed frequency data back into the spatial domain. To ensure stability and consistency in the frequency feature distribution, $BN\left(\cdot\right)$ is applied for normalization.

\subsubsection{Global-Local Attention}
The $GLA\left(\cdot\right)$ mechanism is designed to integrate global context information with local detail features, addressing the loss of fine details in specular highlight removal. The specific implementation is formulated as shown in Eq.~(\ref{eq4}):
\begin{equation}
	GLA\left(f,f_p\right)=SAB\left(CAB\left(f_p\right)\otimes f\right)\otimes f \label{eq4}
\end{equation}
where, $f\in R^{N\times C\cdot E^2\times\frac{H}{E}\times\frac{W}{E}}$ and $ f_p\in R^{N\times C\cdot E^2\times\frac{H}{E}\times\frac{W}{E}}$ represent the input features, and $f_p$ denote the corresponding frequency-domain features obtained through the $FP\left(\cdot\right)$ transformation. Channel Attention Block (CAB) and Spatial Attention Block (SAB)~\cite{RN39} are employed to capture the global distribution of highlight regions and enhance the model's ability to restore local details. The symbol $\otimes$ represents pixel-wise multiplication, enabling effective fusion of the attention-enhanced features.

\subsubsection{Contextual Anchor Attention} 
Traditional $ 3 \times 3 $ convolutions often struggle to capture critical pixel information around specular highlight in high-resolution images, resulting in the loss of fine texture details. This limitation arises from their inherently small receptive field, which prevents effective context modeling in these challenging regions. To overcome this, we introduce $CAA\left(\cdot\right)$~\cite{RN38}, designed to strengthen feature representation in the central highlight regions. CAA combines global average pooling to aggregate broader contextual information with 1D strip convolution, which preserves structural continuity along key spatial directions. This combination enhances the model's focus on crucial regions while maintaining the surrounding contextual integrity. This approach works in tandem with $SW\left(\cdot\right)$, improving the model’s ability to capture contextual information over medium and long distances, ultimately supporting more effective highlight removal.

\subsection{Adaptive Region-Aware Hybrid-Domain Dual Attention Convolutional Network}

OAIBlock effectively captures medium-range and long-range contextual information through a comprehensive attention mechanism, allowing for initial localization and feature enhancement of highlight regions. However, it primarily focuses on coarse highlight identification and enhancement, lacking the ability to model degraded textures within these regions at a fine-grained level. To address this limitation, we propose the Adaptive Region-Aware Hybrid-Domain Dual Attention Convolutional Network (HDDAConv). Positioned between OAIBlock and HDCTransformer, this module enables region-adaptive feature decoupling and reconstruction through a dual-path synergetic architecture. By leveraging both spatial and frequency domain features, HDDAConv enhances detail representation in highlight regions and improves the separation of degraded textures from underlying real features. This leads to more effective highlight removal and better texture restoration.

As shown in Fig.~\ref{Fig1}, HDDAConv first applies a window division mechanism to transform the input feature $Input\in R^{N\times C\times H\times W}$ into a feature of size $N\times C\cdot E^2\times\frac{H}{E}\times\frac{W}{E}$. Rather than employing pixel shifting as in OAIBlock, HDDAConv utilizes the Adaptive Hybrid Domain Dual Attention (AHDDA) mechanism to model spatial and frequency domain features at a fine-grained level. This enables precise capture of details across different scales and domains, improving the separation and restoration of degraded textures in highlight regions. The implementation of AHDDA is detailed in Eq.~(\ref{eq5}), where a cross-domain attention mechanism integrates spatial and frequency domain features. This approach enhances fine details while suppressing irrelevant noise, ensuring a more accurate reconstruction. As a result, HDDAConv effectively preserves the true features of highlight regions while restoring degraded textures with high precision.
\begin{equation}
	AHDDA\left(f,f_p\right)=\beta\cdot CAB\left(f_p\right)\otimes f+\left(1-\beta\right)\cdot SAB\left(f_p\right)\otimes f
	\label{eq5}
\end{equation}
where $f\in R^{N\times C\cdot E^2\times\frac{H}{E}\times\frac{W}{E}}$ and $f_p\in R^{N\times C\cdot E^2\times\frac{H}{E}\times\frac{W}{E}}$ denote the input features, and $f_p$ represents the frequency-domain transformation of $f$ processed by the frequency domain processor $FP\left(\cdot\right)$. The learnable parameter $\beta $ adaptively balances the contributions of channel attention $CAB\left(\cdot\right)$ and spatial attention $SAB\left(\cdot\right)$ to optimize feature.

Capturing dependencies between specular highlight regions and their surrounding features in high-resolution images remains challenging with dual attention alone. To overcome this limitation, HDDAConv develops the pixel bi-directional offset and windowing to extract fine highlight features and reconstruct textures more effectively. Additionally, it incorporates double-path contextual anchor attention (CAA) alongside AHDDA to further enhance feature representation. The detailed implementation is illustrated in Fig.~\ref{Fig1}.

\subsection{Hybrid-Domain Cross-Attention Transformer}
At the bottom of the UNet encoder, the feature map has been progressively refined, including high-level semantic representations. While existing Transformer-based methods, such as DHAN-SHR and UNet-Transformer, operate on raw-resolution or shallow features, they incur high computational costs when processing high-resolution images and are susceptible to redundant information. To enhance computational efficiency and improve global feature perception, we employ the Hybrid-Domain Cross-Attention Transformer (HDCTransformer) at downsampled scales of $ H/4 \times W/4 $ and $ H/8 \times W/8 $. This approach is grounded in two key principles:
\subsubsection{Leveraging Deep Feature Semantics}
As the network encodes the image, feature maps transition from low-level textures to abstract and high-level representations. These deeper features are more effective in capturing global structures and long-range dependencies. Applying attention mechanisms at this stage enhances both information exchange efficiency and accuracy.

\subsubsection{Optimizing Computational Efficiency}

The computational complexity of the self-attention mechanism is proportional to the square of the feature map size, i.e., $O(N^{2})$. By downscaling the feature map to 1/4 or 1/8 of the original resolution, the complexity is reduced to $O(\frac{N^{2}}{16})$. This reduces computational resource consumption while maintaining efficient feature representation.

To improve the accuracy of highlight region identification, we integrate the $FP\left(\cdot\right)$ into the HDCTransformer. This processor refines the query matrix $Q$ by incorporating prior information from the frequency domain. By transforming $Q$ in this way, we establish a complementary relationship with the key K and value V matrices, which remain in the spatial domain. This integration of frequency and spatial information enhances the attention mechanism’s ability to detect highlight regions more effectively. Additionally, the combination of these two domains provides cross-domain calibration for the attention weights used in Eq.~(\ref{eq6}). This leads to improved feature representation of highlight regions. The detailed implementation of the HDCTransformer is outlined below:

\begin{equation}
	HDCTransformer\left(Q,K,V\right) \approx Softmax\left(\frac{{Q\cdot K^T}}{\delta}\right)\cdot V\label{eq6}
\end{equation}
where $K$ and $V$ are obtained by mapping the input feature tensor, defined as $Input\in R^{N\times4C\times\frac{H}{4}\times\frac{W}{4}}$. The $Q$ is generated by applying the frequency domain processor $FP\left(\cdot\right)$ to the same input. Dimensionality reduction decreases memory consumption during training and enhances inference efficiency, while preserving the HDCTransformer’s ability to model features effectively. Next, the resulting features are reshaped into a tensor of shape $N\times Head\times Channel\times H\cdot W$, where $N$ is the batch size, and $Head$ is the number of attention heads, calculated as $dim/16$. This structure enables the model to capture diverse information across multiple subspaces. Each attention head is assigned $dim/Head$ channels, where $dim$ is either $4C$ or $8C$, depending on the configuration. To further enhance flexibility, we replace the standard fixed scaling factor with a learnable parameter $\delta$, initialized to $dim/16$. This parameter is updated through backpropagation, allowing the model to adaptively control the distribution of attention weights. As a result, the model can dynamically adjust the intensity of attention across different image regions, leading to more effective detail restoration in highlight areas.

\subsection{Double Out}
Our model is uniquely designed to output two complementary images in an end-to-end manner: one representing the highlight-free image $Out_1$, and the other the corresponding specular highlight residual $Out_2$. These two outputs sum pixel-wise to reconstruct the original input image $Input$ containing specular highlights, satisfying the following physical constraint Eq.~(\ref{eq7}):
 \begin{equation}
	\begin{aligned}
        Input=\sum_{c=1}^{3} \sum_{i=1}^{H} \sum_{j=1}^{W}Out_1(x_{c,i,j})+Out_2(x_{c,i,j})
	\end{aligned}
	\label{eq7}
\end{equation}
A key challenge lies in the fact that datasets only contain paired highlight-free and highlight images, without explicit ground truth for the specular highlight residual. To address this, we implicitly extract the highlight residual by leveraging the physical property that the input image can be decomposed into a sum of diffuse (highlight-free) and specular (highlight residual) components. By enforcing the reconstruction constraint through the loss function Eq.~(\ref{eq8}):
 \begin{equation}
	\begin{aligned}
		\mathcal{L}=\mathcal{L}_{MSE}(Out_1+Out_2,Input)
	\end{aligned}
	\label{eq8}
\end{equation}
where $\mathcal{L}_{MSE}(\cdot,\cdot)$ denotes a pixel-wise loss function.
This loss encourages the model to learn a meaningful separation between specular and specular components-free despite the absence of explicit supervision on the specular residue. 
By leveraging this end-to-end design, our method jointly models specular removal and residue extraction, enhancing restoration accuracy while maintaining end-to-end differentiability and training stability.

\subsection{Loss Function}
The goal of specular highlight removal is to generate a diffuse image from one that contains specular highlight. This process is a form of image-to-image translation, where the model learns to transform the input image into a more natural, highlight-free version. To achieve this, a loss function is used to quantify the difference between the ground truth and the inference result of specular highlight removal algorithm. It guides the model's training by encouraging adjustments that reduce this difference. An effective loss function should capture differences in pixel values, texture, edges, and high-level visual features to ensure the generated image is perceptually consistent with the ground truth. To address these requirements, we propose a composite loss function that combines three components: Mean Squared Error ($\mathcal{L}_{MSE}$) for pixel-wise accuracy, Structural Similarity Index Measure ($\mathcal{L}_{SSIM}$) for preserving image structure, and a semantic feature loss ($\mathcal{L}_{VGG}$) based on a pretrained neural network to capture higher-level visual characteristics. This combined approach is formalized in Eq.~(\ref{eq9}):
 \begin{equation}
	\begin{aligned}
		\mathcal{L}&=\alpha\cdot\mathcal{L}_{MSE_1}(Out_1,GT) +\beta\cdot\mathcal{L}_{MSE_2}(Out_1+Out_2,Input)\\
		&\quad +\gamma\cdot\mathcal{L}_{SSIM}+\delta\cdot\mathcal{L}_{VGG}
	\end{aligned}
	\label{eq9}
\end{equation}

where the weights $\alpha$, $\beta $, $\gamma$, and $\delta$ are set to 1, 1.5, 0.4, and 0.2, respectively. $Out_1$ refers to the generated image with specular highlight removal, while $Out_2$ represents the version with highlight retained. The term $Input$ denotes the original input image provided to the model. To evaluate high-level perceptual similarity, we use the semantic feature loss ($\mathcal{L}_{VGG}$). 

This loss measures the difference between the generated image and the ground truth by comparing their representations in a feature space extracted using a pretrained VGG16 network. The specific formulation of this loss is provided in Eq.~(\ref{eq10}).
\begin{equation}
	\mathcal{L}_{VGG}=\frac{1}{N}\sum_{l}{\parallel\phi_l\left(\hat{I}\right)-\phi_l\left(I\right)\parallel}_2^2,\ l\in{3,8,15}\label{eq10}
\end{equation}
where $\hat{I}$ represents the image generated by the model, $I$ denotes the corresponding ground truth image, $\phi_l$ is the feature extraction function from the $l$ layer of the VGG network, and $N$ is the total number of layers used in computing the semantic feature loss.

\section{Experiments}
\subsection{Implementation Details}
We implemented the specular highlight removal algorithm, MM-SHR, using the PyTorch framework and conducted all experiments on an Ubuntu 24.04 system. The hardware setup for these experiments included an AMD EPYC 7F72 24 core processor and an NVIDIA GeForce RTX 4090D GPU. For optimization, we used the AdamW optimizer with a batch size of 16. The model was trained for 100 epochs, with cosine annealing for learning rate scheduling. The initial learning rate was set to 1e-3, with a minimum value of 1e-5. All other parameters were set to their default values.

\subsection{Evaluation Metrics} 
To assess the performance of our specular highlight removal method, we selected three widely used image quality assessment metrics: Peak Signal-to-Noise Ratio (PSNR), Structural Similarity Index (SSIM), and Learned Perceptual Image Patch Similarity (LPIPS). PSNR is a traditional metric that measures image fidelity by calculating the ratio between the maximum possible pixel value and the mean squared error. While simple and computationally efficient, it does not always reflect perceived visual quality. SSIM improves on this by evaluating structural similarity between images, taking into account luminance, contrast, and local patterns. It better aligns with how the human visual system perceives image quality. LPIPS offers a more advanced, perceptually driven evaluation. It uses a pre-trained deep neural network to extract high-level features and compare differences at a semantic level, capturing subtle perceptual variations that traditional metrics may miss. Together, these three metrics provide a comprehensive evaluation of our method across multiple dimensions, including pixel-level accuracy, structural preservation, and perceptual consistency, offering a robust, quantitative basis for objective comparison.

\subsection{Comparison with State-of-the-Art Methods}
To comprehensively evaluate the effectiveness of the proposed MM-SHR framework, this paper conducts rigorous quantitative and qualitative comparative experiments on three benchmark datasets for specular highlight removal (SHIQ~\cite{RN43}, PSD~\cite{RN41}, NSH~\cite{RN37}), comparing it with six advanced methods (SLRR~\cite{RN40}, SpecularityNet~\cite{RN41}, TSHRNet~\cite{RN17}, HighlightRNet~\cite{RN37}, DHAN-SHR~\cite{RN34}, AHA~\cite{RN42}). Importantly, the data used in these experiments were not part of the training datasets, ensuring an unbiased assessment. To maintain fairness, all comparison methods were re-trained on our team's hardware devices. We also optimized the hyperparameters based on the settings from the original papers to ensure the best possible performance. Furthermore, the SHIQ, PSD, and NSH datasets do not include ground truth intrinsic images required for the first stage of TSHRNet training. To address this, we adapted the TSHRNet architecture to meet the specific needs of this experiment.
\begin{table*}[!t]
	\caption{ Quantitative comparison of MM-SHR with six advanced specular highlight removal methods on SHIQ, PSD, and NSH. The top-performing results are highlighted in bold red, the second-best results in bold black, and the third-best results are underlined.}
	\label{table1}
	\begin{center}
		\begin{tabular}{ccccccccccccc}
			\toprule 
			&\multicolumn{3}{c}{SHIQ} 
            &\multicolumn{3}{c}{PSD} & \multicolumn{3}{c}{NSH} & \multicolumn{3}{c}{Average} \\
			\cmidrule(lr){2-4}
            \cmidrule(lr){5-7}
            \cmidrule(lr){8-10}
            \cmidrule(lr){11-13}
            \multirow{-2}{*}{Methods}           
            & PSNR↑                                   & SSIM↑                                  & LPIPS↓                                 & PSNR↑                                   & SSIM↑                                  & LPIPS↓                                 & PSNR↑                                   & SSIM↑                                 & LPIPS↓                                & PSNR↑                                   & SSIM↑                                  & LPIPS↓                                 \\
			\hline
			SLRR~\cite{RN40}                                 & 14.6890                                  & 0.6942                                 & 0.2703                                 & 13.3136                                 & 0.5329                                 & 0.2323                                 & 15.5497                                 & 0.5621                                & 0.2137                                & 14.5174                                 & 0.5964                                 & 0.2388                                 \\
			SpecularityNet~\cite{RN41}                       & 32.1849                                 & 0.9581                                 & \textbf{0.0456}                        & 25.2063                                 & {\ul 0.8920}                            & {\color[HTML]{C00000} \textbf{0.0423}} & 40.5315                                 & 0.0059                                & 0.0059                                & 32.6409                                 & 0.6187                                 & \textbf{0.0313}                        \\
			TSHRNet~\cite{RN17}                              & 32.7168                                 & 0.9681                                 & 0.0533                                 & 22.0323                                 & 0.8260                                  & 0.0782                                 & 36.3057                                 & 0.9571                                & 0.0176                                & 30.3516                                 & 0.9171                                 & 0.0497                                 \\
			HighlightRNet~\cite{RN37}                        & 30.9728                                 & 0.9493                                 & 0.0695                                 & \textbf{25.7563}                        & 0.8916                                 & 0.0476                                 & 39.6003                                 & {\ul 0.9765}                          & 0.0067                                & 32.1098                                 & 0.9391                                 & 0.0413                                 \\
			DHAN-SHR~\cite{RN34}                             & \textbf{34.6894}                        & \textbf{0.9774}                        & 0.3667                                 & 25.1358                                 & 0.8918                                 & {\ul 0.0434}                           & {\ul 39.9630}                            & 0.9745                                & \textbf{0.0062}                       & {\ul 33.2627}                           & {\ul 0.9479}                           & 0.1388                                 \\
			AHA~\cite{RN42}                                  & {\ul 34.1620}                            & {\ul 0.9699}                           & {\ul 0.0475}                           & {\ul 25.2260}                            & \textbf{0.8979}                        & 0.0477                                 & \textbf{41.8759}                        & \textbf{0.9861}                       & {\ul 0.0064}                          & \textbf{33.7546}                        & \textbf{0.9513}                        & {\ul 0.0339}                           \\
			{\color[HTML]{C00000} \textbf{MM-SHR(Ours)}} & {\color[HTML]{C00000} \textbf{36.6312}} & {\color[HTML]{C00000} \textbf{0.9808}} & {\color[HTML]{C00000} \textbf{0.0274}} & {\color[HTML]{C00000} \textbf{25.7638}} & {\color[HTML]{C00000} \textbf{0.9034}} & \textbf{0.0433}                        & {\color[HTML]{C00000} \textbf{45.7816}} & {\color[HTML]{C00000} \textbf{0.9940}} & {\color[HTML]{C00000} \textbf{0.0020}} & {\color[HTML]{C00000} \textbf{36.0588}} & {\color[HTML]{C00000} \textbf{0.9594}} & {\color[HTML]{C00000} \textbf{0.0242}}\\
			\bottomrule 
		\end{tabular}
	\end{center}
\end{table*}

\subsubsection{Quantitative Comparison}
Table~\ref{table1} presents a quantitative comparison of MM-SHR with six state-of-the-art specular highlight removal methods. The results consistently demonstrate MM-SHR's superior performance across various datasets. In the SHIQ dataset, MM-SHR achieves a PSNR of 36.6312 dB, outperforming the next best method, DHAN-SHR, by 1.9418 dB (34.6894 dB). MM-SHR also achieves the highest SSIM (0.9808) and the lowest LPIPS (0.0274). Notably, DHAN-SHR's LPIPS on SHIQ is significantly higher at 0.3667, likely due to its model's sensitivity to complex reflectance variations, which leads to a loss of image details. This further emphasizes MM-SHR's ability to preserve image structure. In the NSH dataset, MM-SHR excels with a PSNR of 45.7816 dB, a 3.9057 dB improvement over the second-best method, AHA (41.8759 dB). MM-SHR also achieves a remarkable SSIM of 0.9940 and a minimal LPIPS of 0.0020. While MM-SHR shows strong performance in the PSD dataset as well, with a PSNR of 25.638 dB and an SSIM of 0.9034, its LPIPS (0.0433) is slightly higher than that of SpecialarityNet (0.0423). This small difference is likely due to limitations within the PSD dataset, including misalignment of image pairs, luminance discrepancies, and non-one-to-one correspondence, which can affect highlight removal. Despite this, MM-SHR's balanced performance in PSNR and SSIM, combined with its clear advantages on the SHIQ and NSH datasets, highlight its robust generalization capability. Its superior performance across all three datasets compared to the six other methods underscores MM-SHR's effectiveness and generalization capability in specular highlight removal.
Furthermore, the computational cost of MM-SHR is analyzed, with the results presented in Table ~\ref{table2}.

\begin{table}[!h]
	\setlength{\tabcolsep}{3pt}
	\caption{A comparative analysis of FLOPs and Parameters with state-of-the-art methods, where the input feature size is fixed at $F\in R^{1\times 3\times 256\times 256}$}
	\label{table2}
	\begin{center}
		\begin{tabular}{ccccccc}
			\toprule 
			Methods&        FLOPs(G)↓                    &                       Params(M)↓ \\
			\hline
                SpecularityNet~\cite{RN41} & 212.87 & {\ul17.00} \\
                TSHRNet~\cite{RN17} & {\ul72.76} & 116.99 \\
                HighlightRNet~\cite{RN37} & 250.84 & 48.37 \\
                DHAN-SHR~\cite{RN34} & \textbf{65.69} & {\color[HTML]{C00000} \textbf{4.53}} \\
                AHA~\cite{RN42} & 92.40  & 35.86 \\
                {\color[HTML]{C00000} \textbf{MM-SHR(Ours)}} &
                {\color[HTML]{C00000} \textbf{18.03}} &
                \textbf{16.09} \\
			\bottomrule 
		\end{tabular}
	\end{center}
\end{table}

\subsubsection{Qualitative Comparison}
\begin{figure*}[!t]
 	\includegraphics[width=\linewidth]{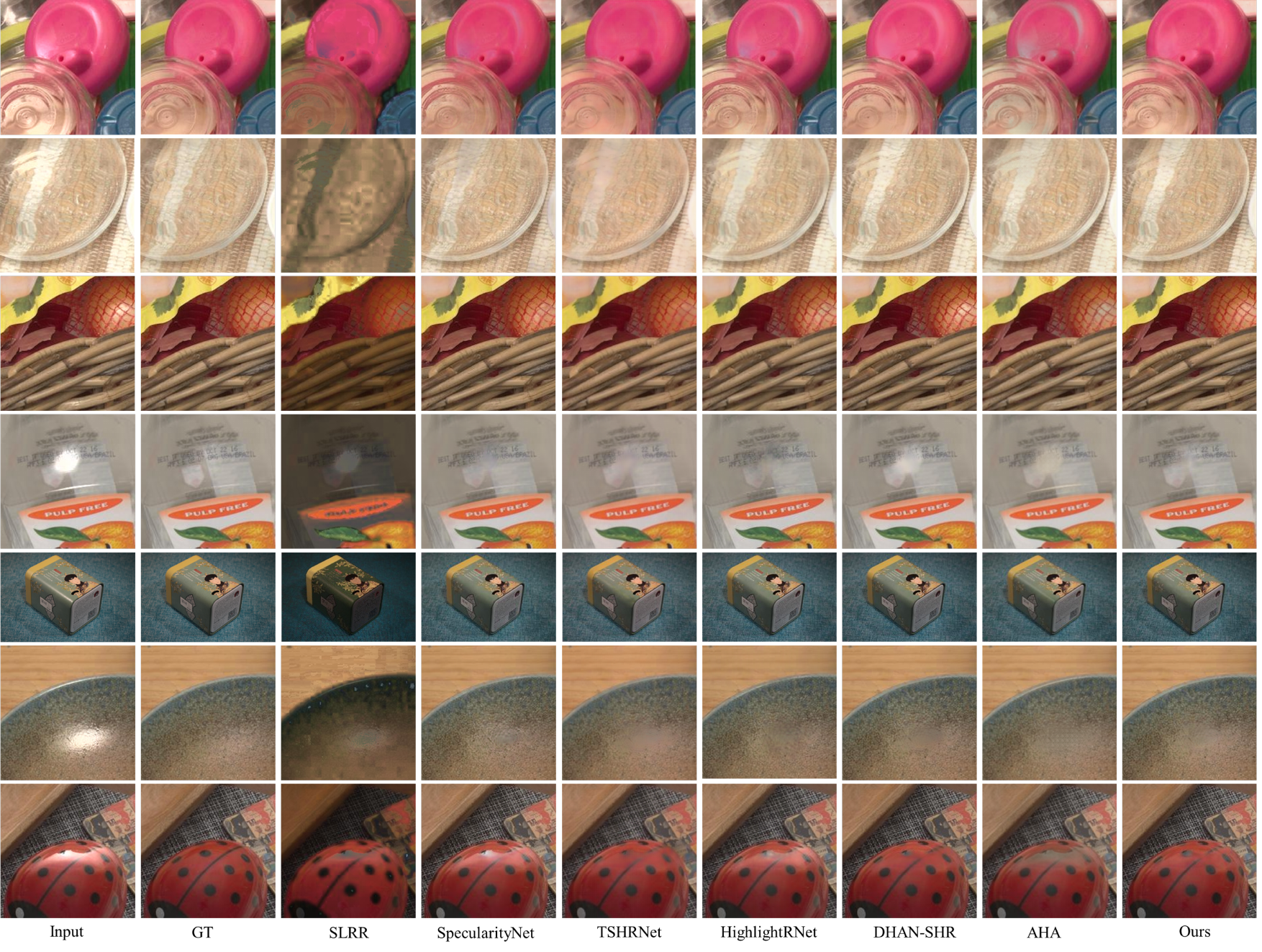}
 	\caption{Qualitative comparison of MM-SHR with leading state-of-the-art highlight removal methods.}
 	\label{Fig3}
\end{figure*}

\begin{figure}[!t]
 	\includegraphics[width=\linewidth]{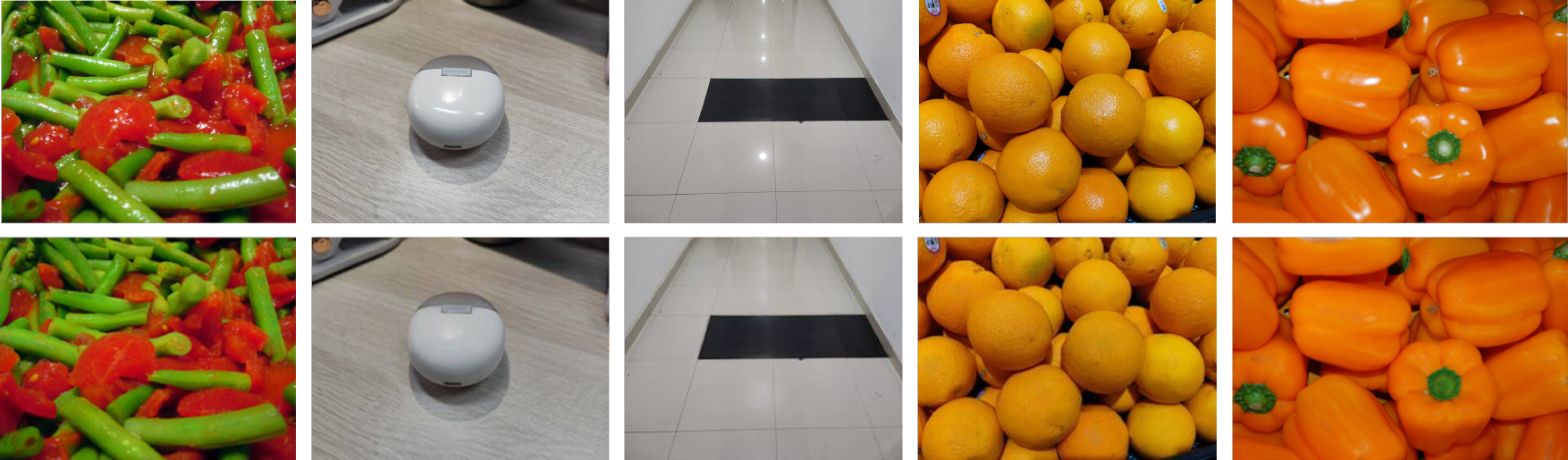}
 	\caption{Zero-shot test results demonstrating the generalization capability of MM-SHR.}
 	\label{Fig4}
\end{figure}

 Fig.~\ref{Fig3} presents a qualitative comparison among MM-SHR and six state-of-the-art methods for specular highlight removal. For a more detailed evaluation, we recommend enlarging the image. The figure demonstrates that, while previous methods achieve some level of success in removing specular highlight, they still exhibit notable limitations. For example, the SLRR method causes overall image darkening and leaves highlight residues. SpecularityNet struggles with significant highlight remnants, particularly on highly saturated materials (see the third and seventh rows). TSHRNet tends to distort the original image structure and texture (see the fourth and sixth rows). Both HighlightRNet and AHA show poor performance on transparent materials (see the second row), with HighlightRNet displaying more prominent highlight residues (see the first and seventh rows) and AHA suffering from severe visual artifacts (see the sixth row) and color distortion (see the seventh row). In contrast, our method, MM-SHR, provides superior visual results. Although DHAN-SHR also delivers strong visual effects, its use of a Transformer module leads to slower inference times, making it less suitable for practical applications. To further demonstrate MM-SHR's robustness and consistency in preserving visual texture,  Fig.~\ref{Fig4} shows the results of Zero-Shot tests. These results clearly highlight MM-SHR's ability to effectively remove specular highlight and its strong generalization capability.

\subsection{Ablation Studies}
To thoroughly evaluate the contribution of each key module and loss function in MM-SHR, we conducted a systematic ablation study on the SHIQ dataset. This experiment assesses how different module combinations affect three metrics: PSNR, SSIM, and LPIPS.

\subsubsection{Fundamental module}
To examine the effectiveness of each component, we designed two sets of comparison experiments under the same conditions of parameters and training methods. In the first set, we began with a baseline UNet architecture and progressively added key modules such as the OAIBlock and HDDAConv. After each addition, the network was retrained and evaluated. The results, shown in Table~\ref{table3}, clearly demonstrate the performance improvements contributed by each module. In the second set of experiments, we started with the full MM-SHR architecture and selectively removed one module at a time, keeping all other components unchanged. The network was then retrained and evaluated. The results, summarized in Table~\ref{table4}, highlight the performance degradation caused by the absence of each individual module. Taken together, these results confirm that each module contributes positively to the overall performance. When fully integrated, MM-SHR achieves the highest performance across all three metrics, indicating that the architecture is well-structured and that the modules work in synergy to enhance highlight removal.

\begin{table*}[!t]
	\caption{Ablation results of MM-SHR on the SHIQ dataset.}
	\label{table3}
	\begin{center}
		\begin{tabular}{cccccccc}
			\toprule 
			&                            &                               &                                  &                              & \multicolumn{3}{c}{SHIQ}                                                                                                  \\
			\cline{6-8}
			\multirow{-2}{*}{Unet} & \multirow{-2}{*}{OAIBlock} & \multirow{-2}{*}{HDDAConv} & \multirow{-2}{*}{HDCTransformer} & \multirow{-2}{*}{Double Out} & PSNR↑                                   & SSIM↑                                  & LPIPS↓                                 \\
			\hline
			\textbf{\checkmark}             & \textbf{}                  & \textbf{}                     & \textbf{}                        & \textbf{}                    & 33.4811                                 & 0.9742                                 & 0.0388                                 \\
			\textbf{\checkmark}             & \textbf{\checkmark}                 & \textbf{}                     & \textbf{}                        & \textbf{}                    & 34.1417                                 & 0.9757                                 & 0.0368                                 \\
			\textbf{\checkmark}             & \textbf{}                  & \textbf{\checkmark}                    & \textbf{}                        & \textbf{}                    & 33.6888                                 & 0.9743                                 & 0.0405                                 \\
			\textbf{\checkmark}             & \textbf{\checkmark}                 & \textbf{\checkmark}                    & \textbf{}                        & \textbf{}                    & 34.5779                                 & 0.9764                                 & 0.0368                                 \\
			\textbf{\checkmark}             & \textbf{\checkmark}                 & \textbf{\checkmark}                    & \textbf{\checkmark}                       & \textbf{}                    & 36.4637                                 & 0.9804                                 & 0.0283                                 \\
			\textbf{\checkmark}             & \textbf{\checkmark}                 & \textbf{\checkmark}                    & \textbf{\checkmark}                       & \textbf{\checkmark}                   & {\color[HTML]{C00000} \textbf{36.6312}} & {\color[HTML]{C00000} \textbf{0.9808}} & {\color[HTML]{C00000} \textbf{0.0274}}\\
			\bottomrule 
		\end{tabular}
	\end{center}
\end{table*}

\begin{table}[!t]
	\caption{Ablation results of MM-SHR individual component on the SHIQ dataset.}
	\label{table4}
	\begin{center}
		\begin{tabular}{cccc}
			\toprule 
            & \multicolumn{3}{c}{SHIQ}  \\
            \cline{2-4}
			\multirow{-2}{*}{Without} & PSNR↑                                   & SSIM↑                                  & LPIPS↓                                 \\
            \hline
			OAIBlock                  & 36.5231                                 & 0.9807                                 & 0.0279                                 \\
			HDDAConv               & 35.8800                                   & 0.9755                                 & 0.0372                                 \\
			HDCTransformer            & 33.3801                                 & 0.9678                                 & 0.0479                                 \\
			all                       & {\color[HTML]{C00000} \textbf{36.6312}} & {\color[HTML]{C00000} \textbf{0.9808}} & {\color[HTML]{C00000} \textbf{0.0274}}\\
			\bottomrule 
		\end{tabular}
	\end{center}
\end{table}

\subsubsection{Loss Function}
To assess the effectiveness of our composite loss function, we conducted a series of experiments by incrementally introducing additional loss components $\mathcal{L}_{MSE_2}$, $\mathcal{L}_{SSIM}$, and $\mathcal{L}_{VGG}$ on top of $\mathcal{L}_{MSE_1}$. Each loss term was assigned an appropriate weight to evaluate its individual contribution to highlight removal performance. As shown in Table~\ref{table5}, MM-SHR achieves its best results only when all loss components are used together. This confirms the effectiveness and necessity of the composite loss function design.

\begin{table}[!t]
	\setlength{\tabcolsep}{3pt}
	\caption{Ablation experiments for MM-SHR loss function on the SHIQ dataset.}
	\label{table5}
	\begin{center}
		\begin{tabular}{ccccccc}
			\toprule 
			&                           &                           &                          & \multicolumn{3}{c}{SHIQ}                                                                                                  \\
			\cline{5-7}
			\multirow{-2}{*}{$\mathcal{L}_{MSE_1}$} & \multirow{-2}{*}{$\mathcal{L}_{MSE_2}$} & \multirow{-2}{*}{$\mathcal{L}_{SSIM}$} & \multirow{-2}{*}{$\mathcal{L}_{VGG}$} & PSNR↑                                   & SSIM↑                                  & LPIPS↓                                 \\
			\hline
			\textbf{\checkmark}                        &                           &                           &                          & 34.944                                  & 0.9736                                 & 0.0400                                   \\
			\textbf{\checkmark}                        & \textbf{\checkmark}                        &                           &                          & 35.8238                                 & 0.9758                                 & 0.0377                                 \\
			\textbf{\checkmark}                        & \textbf{\checkmark}                        & \textbf{\checkmark}                        &                          & 36.6242                                 & 0.9806                                 & 0.0321                                 \\
			\textbf{\checkmark}                        & \textbf{\checkmark}                        & \textbf{\checkmark}                        & \textbf{\checkmark}                       & {\color[HTML]{C00000} \textbf{36.6312}} & {\color[HTML]{C00000} \textbf{0.9808}} & {\color[HTML]{C00000} \textbf{0.0274}}\\
			\bottomrule 
		\end{tabular}
	\end{center}
\end{table}

\section{Conclusion}
We propose MM-SHR, a multi-model cooperative framework, to address two key challenges in specular highlight removal: the preservation of fine texture details and the balance between accuracy and efficiency. The framework adopts a hierarchical architecture that combines convolutional layers for detailed feature extraction with Transformer-based modules for global context integration, enabling both high-quality and efficient specular highlight removal. To enhance feature representation, we introduce two novel modules, OAIBlock and HDDAConv, based on frequency domain-guided, pixel shifting, and windowing mechanisms. These modules facilitate the effective coupling of high-resolution medium-range and long-range contextual features in the top layers of the network. In parallel, the HDCTransformer module is incorporated at the bottleneck layer to capture abstract global features, thereby improving the overall modeling capacity and accuracy of highlight suppression. Extensive evaluations on three benchmark datasets demonstrate that MM-SHR achieves strong performance and fast inference speed. Systematic ablation studies confirm the effectiveness and complementary roles of each module within the architecture. 

\bibliographystyle{IEEEtran}
\bibliography{references}
\end{document}